\documentclass[pdflatex,sn-mathphys]{sn-jnl}

\jyear{2021}%

\theoremstyle{thmstyleone}%
\usepackage{amssymb}
\usepackage{booktabs,xltabular}
\usepackage{float}
\usepackage{titlesec} 
\usepackage{tabularx}
\usepackage{multirow}
\theoremstyle{thmstyletwo}%
\newtheorem{definition}{Definition}

\theoremstyle{thmstylethree}%

\begin{document}

\title[Article Title]{Benchmarks for Pirá 2.0, a Reading Comprehension Dataset about the Ocean, the Brazilian Coast, and Climate Change}


\author*[1]{\fnm{Paulo} \sur{Pirozelli}}\email{paulo.pirozelli.silva@usp.br}

\author[4]{\fnm{Marcos} \sur{M. José}}

\author[2]{\fnm{Igor} \sur{Silveira}}

\author[4]{\fnm{Flávio} \sur{Nakasato}}

\author[3]{\fnm{Sarajane} \sur{M. Peres}}

\author[4]{\fnm{Anarosa} \sur{A. F. Brandão}}

\author[4]{\fnm{Anna} \sur{H. R. Costa}}

\author[4]{\fnm{Fabio} \sur{G. Cozman}}

\affil[1]{\orgdiv{Instituto de Estudos Avançados}}


\affil[2]{\orgdiv{Instituto de Matemática e Estatística}}

\affil[3]{\orgdiv{Escola de Artes, Ciências e Humanidades}}

\affil[4]{\orgdiv{Escola Politécnica}}


\affil{Universidade de São Paulo}


\abstract{Pirá is a reading comprehension dataset focused on the ocean, the Brazilian coast, and climate change, built from a collection of scientific abstracts and reports on these topics. This dataset represents a versatile language resource, particularly useful for testing the ability of current machine learning models to acquire expert scientific knowledge. Despite its potential, a detailed set of baselines has not yet been developed for Pirá. By creating these baselines, researchers can more easily utilize Pirá as a resource for testing machine learning models across a wide range of question answering tasks. In this paper, we define six benchmarks over the Pirá dataset, covering closed generative question answering, machine reading comprehension, information retrieval, open question answering, answer triggering, and multiple choice question answering. As part of this effort, we have also produced a curated version of the original dataset, where we fixed a number of grammar issues, repetitions, and other shortcomings. Furthermore, the dataset has been extended in several new directions, so as to face the aforementioned benchmarks: translation of supporting texts from English into Portuguese, classification labels for answerability, automatic paraphrases of questions and answers, and multiple choice candidates. The results described in this paper provide several points of reference for researchers interested in exploring the challenges provided by the Pirá dataset.}




\maketitle

\section{Introduction}\label{sec1}


The performance of machine learning models is strongly influenced by the quality of the underlying dataset. Nonetheless, the mere existence of a proper dataset may not be sufficient to guarantee the desired performance. To start, resources may not be easily available to users. Even when a dataset is within reach, researchers may be discouraged to use it  in the absence of established benchmarks and baselines, as it is not clear in this case what should be expected from its use. A good set of benchmarks can also suggest non-trivial uses for a dataset, sometimes in tasks far from its original and intended purpose.


In this paper we deal with Pirá \cite{Paschoal2021}, a dataset about the ocean, the Brazilian coast, and climate change. Pirá is a reading comprehension dataset containing texts, questions and answers in two languages (Portuguese and English), manual paraphrases, and human evaluations. 
It offers a rich tool with which to explore language tasks in general and ocean-related question answering (QA) in particular. 


A dataset such as this, with bilingual questions and answers, supporting texts, and human evaluations, can be explored in countless ways. Thus, in order to set the limits of our investigation, we select six chief benchmarks to pursue: closed generative question answering, machine reading comprehension, information retrieval, open
question answering, answer triggering, and multiple choice question answering. For each benchmark, we establish a number of baselines, including human (when available), random, and machine learning models, giving preference to solutions most commonly used in the corresponding tasks and cutting-edge large language models (LLMs).

\begin{figure*}[tp]
  \centering
  \includegraphics[width=1\textwidth]{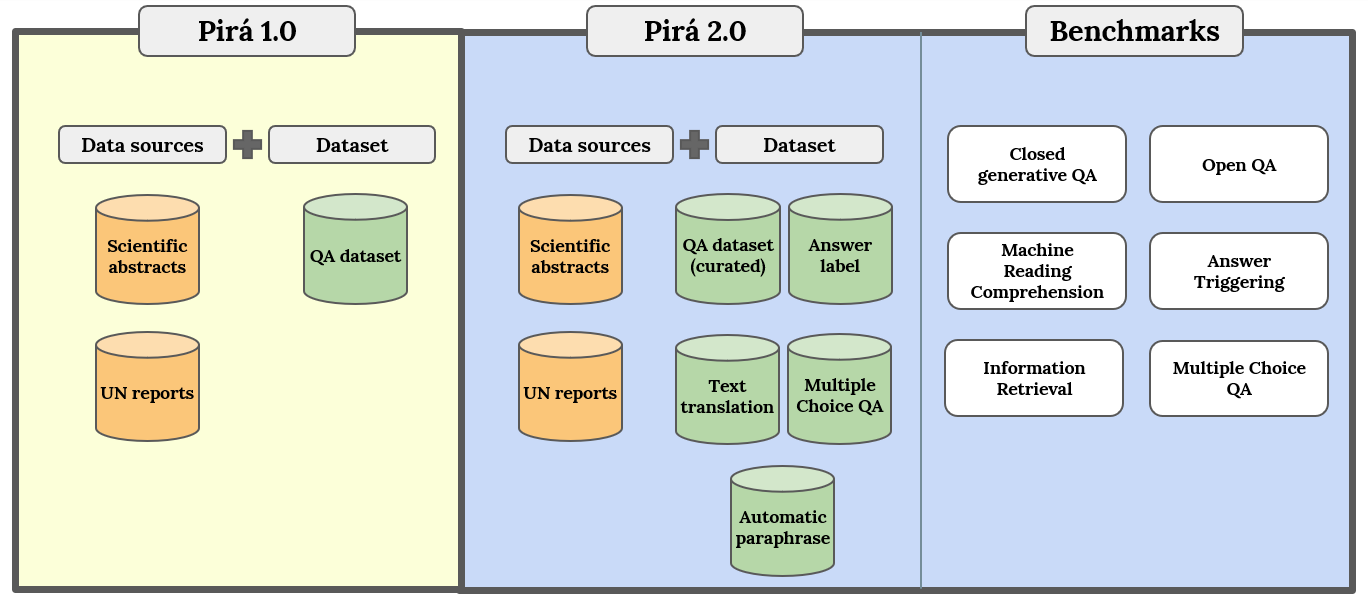}
  \caption{Comparative overview of Pirá 1.0's original structure (yellow background) and the contributions presented in this paper (blue background), divided by dataset modifications and benchmarks. Pirá 2.0 is a curated version of Pirá 1.0 that offers a number of extensions (translations of supporting texts into Portuguese, answerability labels, automatic paraphrases, and multiple choice candidates). Furthermore, six benchmarks are defined for this curated dataset: closed generative question answering, machine reading comprehension, information retrieval, open question answering, answer triggering, and multiple choice question answering. QA stands for ``Question Answering'' and UN for ``United Nations''.}
  \label{fig:pira1_v_pira2}
\end{figure*}

As part of our enterprise, we also introduce a curated version of the original dataset, which we refer to as \textbf{Pirá 2.0}. A complete revision of the original dataset (henceforth, Pirá 1.0), Pirá 2.0 corrects for grammar issues, misplaced entries, repeated questions, and other minor flaws. We also added a number of resources to the original dataset and extended it in several directions, allowing one to run new tasks with it.
Figure \ref{fig:pira1_v_pira2} provides a comparative overview of Pirá 1.0' original structure and the contributions presented in this paper, separated by dataset modifications
and benchmarks.
 
Every experiment  described in this paper has been run on Pirá 2.0, which is available to download from our GitHub page;\footnote{\url{https://github.com/C4AI/Pira}} codes for tests can be found in the same page. We also include a leaderboard in our GitHub repository, where the main results for each benchmark are displayed. 

The paper is organized as follows. Section \ref{sec:dataset} describes the original Pirá 1.0 dataset as well as the improvements and additions made so as to obtain Pirá 2.0. The six benchmarks are then explored in the following sections: closed generative question answering (Section \ref{sec:no_context}), machine reading comprehension (Section \ref{sec:mrc}), information retrieval (Section \ref{sec:ir}), open question answering (Section \ref{sec:oqa}), answer triggering (Section \ref{sec:AT}), and multiple choice question answering (Section \ref{sec:mcqa}). These sections are relatively self-contained and can be read independently. Sections \ref{sec:no_context} to \ref{sec:oqa} present standard tasks on natural language processing (NLP) and for this reason are more focused on results; Sections \ref{sec:AT} and \ref{sec:mcqa} spend relatively more time on defining the benchmarks. Section \ref{sec:discussion} discusses the baselines from the perspective of the models' size, which goes from 110M for BERT base up to 175B for GPT3-turbo (and possibly more for GPT4). In Section \ref{sec:limitations}, we consider some limitations of Pirá 2.0 and future paths for extension and improvement. Finally, in Section \ref{sec:conclusion} we discuss the main insights gathered from these benchmarks and point to the challenges still posed by Pirá 2.0.

\section{The Dataset} \label{sec:dataset}

We first summarize, in Section \ref{subsection:Pira1}, the main features of Pirá 1.0, a dataset produced by our group and previously described in a conference paper \cite{Paschoal2021}.
We then describe, in Section \ref{subsection:Pira2}, the improvements and additions made for Pirá 2.0.

\subsection{Pirá 1.0}\label{subsection:Pira1}

Pirá 1.0 is a bilingual reading comprehension dataset about the ocean, the Brazilian coast, and climate change. The supporting texts used to generate questions were taken from the scientific literature, making this a dataset proper for evaluating expert and scientific knowledge. Each instance of Pirá 1.0 dataset comprises up to nine textual elements: supporting text, original question (PT-EN),\footnote{PT-EN stands for Portuguese and English, respectively.} original answer (PT-EN), validation answer (PT-EN), paraphrased question (PT-EN), and up to five types of evaluation. In total, the dataset contains 2261 of these QA sets.\footnote{We refer to the multiple elements that comprise a single observation as a question answering (QA) set.}



The construction of Pirá 1.0 was structured around four main phases.\footnote{For further details about the dataset construction process and statistics, we refer to the original paper \cite{Paschoal2021}.} The majority of the work was done by a group of annotators, undergraduate and graduate students of information systems and computer engineering. 

\begin{enumerate}
    \item \textit{Corpus Collection.} Two corpora of ocean-related texts were collected by the authors of the paper. The first corpus consisted of abstracts of papers about the Brazilian coast obtained from a scientific article database (Scopus database), 
    and the second consisted of small excerpts of two United Nations reports about the ocean \cite{un2017world, un2021world}.
    \item \textit{QA Creation.} Annotators used the supporting texts to produce questions and answers in Portuguese and English (the same question expressed in the two languages). They were allowed to use automatic translation tools for this task, but were required to check the resulting translations. A web application was specially developed to manage the activity. For the corpus of scientific abstracts, annotators had to decide whether the abstract was related to the subject (i.e., Brazilian maritime territory); supporting texts that were unrelated to the subject were excluded from the database.
    \item \textit{Partial Editing.} The authors manually edited the questions and answers for spelling and grammar. They also removed questions that were excessively decontextualized (e.g., ``What is the aim of the paper?'').
    \item \textit{Assessment.} Annotators assessed the material produced in the QA creation phase. More than 90\% of the QA sets were fully assessed in this way. The process involved three different tasks, which took place in the same web application. 
    \begin{enumerate}
        \item \textit{Validation.} Annotators were asked to answer the questions made in the QA creation phase. We refer to these answers throughout the paper as \textit{validation answers}.
        \item \textit{Evaluation.} Annotators also evaluated the original QA sets according to a number of aspects.\footnote{The five evaluations consisted of (a) two yes/no questions, (b) the evaluation of three statements according to a Likert scale (1 - Strongly disagree, 2 - Disagree, 3 - Neither agree nor disagree, 4 - Agree, 5 - Strongly agree) and a classification of the question type: who, what, where, when, why, how, none of them. The yes/no questions comprise: (1) Is this a generic question? Example: ``How big is the oil field?''; (2) Can you answer the question only with the information provided by the text?. The scaled statetements comprise:  (1) The question makes sense; (2) The question is hard; (3) Your answer and the original answer are equivalent.}
        \item \textit{Question Paraphrasing.} Annotators were asked to paraphrase the original questions. 
    \end{enumerate}
    
\end{enumerate}

Table \ref{tab:QA_example} depicts an example of a typical QA set from the Pirá 1.0 dataset.

\begin{table}[htp]
  \caption{Example of QA set from Pirá 1.0 [ID: B2335]. Supporting text has been truncated for expository purposes.}
  \label{tab:QA_example}
  \begin{tabular}{p{3.65cm}|p{4cm}|p{3cm}}
    \hline
    \textbf{Supporting text} & \textbf{Textual elements} & \textbf{Evaluations}\\
    \hline
    {\footnotesize […] The areas supporting high relative and absolute levels of biodiversity not only harbour unique species adapted to their special features, but also often serve as centres for essential life-history stages of species with wider distributions. For example, essentially all the biodiversity hotspots that have been identified have also been found to harbour juvenile fish, which are important for fisheries in adjacent areas. Hotspots for primary productivity are necessarily also hotspots for production of oxygen as a direct result of photosynthesis. Furthermore, underlying the high biodiversity is often a high structural complexity of the habitats that support it. That structure often contributes other services, such as coastal protection and regeneration. In addition, it is the concentrated presence of iconic species in an area which adds to aesthetic services (supporting tourism and recreation) and spiritual and cultural services.} & {\footnotesize \textbf{Question (EN):} What is the importance of ocean regions with high biodiversity?} \newline {\footnotesize \textbf{Question (PT):} Qual é a importância de regiões oceânicas com alta biodiversidade?} \newline {\footnotesize \textbf{Answer (EN):} They support fishing in adjacent areas, oxygen production, coastal protection and regeneration, tourism and recreation, and spiritual and cultural services.} \newline {\footnotesize \textbf{Answer (PT):} Elas favorecem a pesca em áreas adjacentes, produção de oxigênio, proteção e regeneração costeira, turismo e recreação, e atividades culturais e espirituais.} \newline {\footnotesize \textbf{Validation (EN):} Serve as centers for essential stages in the life history of species with wider distribution.} \newline {\footnotesize \textbf{Validation (PT):} Servir como centros para estágios essenciais da história de vida de espécies com distribuição mais ampla.} \newline {\footnotesize \textbf{Paraphrase (EN):} What is the relevance of oceanic regions with high biodiversity?} \newline {\footnotesize \textbf{Paraphrase (PT):} Qual a relevância das regiões oceânicas com alta biodiversidade?} & {\footnotesize \textbf{Is this a generic question? Example: “How big is the oil field?”:} Yes} \newline
    {\footnotesize \textbf{Can you answer the question with just the information provided by the text?:} Yes} \newline {\footnotesize \textbf{The question makes sense:} Strongly agree (5)} \newline {\footnotesize \textbf{The question is hard:} Neither agree nor disagree (3)} \newline {\footnotesize \textbf{Your answer and the original answer are equivalent:} Strongly disagree (1)} \newline {\footnotesize \textbf{Type of question:} What}\\ 
    \hline
  \end{tabular}
\end{table}

\subsection{Pirá 2.0}\label{subsection:Pira2}

Pirá 1.0 left many opportunities  for extension and improvement. To start, the QA sets presented many grammar issues (especially in the English part of the dataset, as that was not the participants'  first language). We also curated the dataset for minor formating issues, such as: exchanged cells between Portuguese and English; incorrect line breaks; problems with non-Latin characters; absence of question IDs; and repeated entries. 

For this reason, we conducted a thorough revision of the dataset. We reviewed and edited the QA sets, except for the original questions and answers, as those had already been edited during the \textit{Partial Editing} phase; moreover, altering them would have resulted in inconsistencies with the evaluations. As part of this editing process, we also incorporated questions and evaluations that were produced after the release of Pirá 1.0, resulting in a similar number of QA sets (2258 sets) as the original dataset. This curated version of the dataset is referred to as Pirá 2.0. 


More importantly, Pirá 2.0 contains a number of features that are required for the benchmarks discussed in this paper. There are four main extensions introduced in Pirá 2.0:

\begin{enumerate}
    \item \textit{Automatic translations of supporting texts into Portuguese}, allowing us to run the benchmarks for machine reading comprehension in this language as well (Section \ref{sec:mrc});
    \item \textit{Classification labels} indicating whether a question can be answered or not, which are essential to the answer triggering benchmark (Section \ref{sec:AT});
    \item \textit{Automatically-generated paraphrases} for questions and answers in both Portuguese and English, a valuable data augmentation resource for training larger models (Section \ref{sec:AT});    
    \item \textit{Multiple-choice QA extension}, where each question has five candidate answers, of which only one is correct (Section \ref{sec:mcqa}).

\end{enumerate}

Examples of automatic paraphrases and multiple-choice questions can be found in Tables \ref{tab:paraphrase} and \ref{tab:mcqa_example}, respectively. These and the other new features introduced by Pirá 2.0 are explained in more detail in the respective sections.

\begin{table}[htp]
  \caption{Example of automatic paraphrases. Paraphrases in English are produced with PEGASUS; those in Portuguese are made with PTT5-Paraphraser. Square brackets indicate the QA sets' IDs.
  }
  \label{tab:paraphrase}
  \begin{tabular}{p{5.5cm} | p{5.5cm}}
    \hline
    \textbf{Manually-created QAs} & \textbf{Automatic paraphrase}\\
    \hline
    \textbf{Question:} When did the first phase of the development plan conceived for Albacora field start? [ID A775]    & When did the first phase of the program, which was intended for Albacora field, begin? \newline
     \\ 
    \textbf{Answer:} Mature regions, such as the North Sea and the Gulf of Mexico. [ID B2062] &
    Mature areas such as the North Sea or the Gulf of Mexico are all covered. \newline \newline
    \\ 
    \textbf{Question:} Onde podemos achar locais de nascimento de C. limbatus? [ID A1217] & Onde se pode encontrar a localização da família de C. limbatus? \newline
    \\
    \textbf{Answer:} Devido à magnitude da propagação vertical de energia e duração dos eventos. [ID A1942] & Isso é devido à intensidade da transmissão vertical de energia e à duração dos eventos.\\
    \hline
  \end{tabular}
\end{table}

\begin{table}[htp]
  \caption{Example of multiple-choice question [ID: A1069]. Supporting text has been truncated for expository purposes. Correct alternative is marked in bold.}
  \label{tab:mcqa_example}
  \begin{tabular}{p{3.65cm} | p{2.9cm} | p{4cm}}
    \hline
    \small{\textbf{Context}} & \small{\textbf{Question}} & \small{\textbf{Alternatives}}\\
    \hline
    \footnotesize
    By taking a bold step forward in developing the Garoupa field offshore Brazil with subsea techniques, Petroleo Brasileiro SA (Petrobras) will benefit from early production to help satisfy the country's growing energy demand. This article describes the first phase of the development program. Petrobras is seeking a producing rate of 45,000 b/d from nine drilled wells. [...] & \footnotesize
How will the Brazilian population be benefited with the early production of petroleum?     &  \footnotesize
    a) \textbf{It will help to supply the growing energy demand}. \newline
b) With submarine flow lines. \newline
c) The provisional system used the seabed pipeline, meter, loading tower and a processing vessel from the original production system. \newline
d) It is a solution for the sand production problems associated with the production of hydrocarbons from sandstone reservoirs. \newline
e) A set of case studies of three typical types of convective systems occurring in Amazonia-i.e., locally occurring systems, coastal-occurring systems and basin-occurring systems.
    \\
    \hline
  \end{tabular}
\end{table}



To run the experiments described in this paper, the dataset was split into three random partitions: training, development, and test sets. Table \ref{tab:dataset_division} presents the number of QA sets for each of these partitions.\footnote{In order to ensure reproducibility, the dataset partitions, as well as the code used to create the split, are made available in the project's GitHub repository.} In all benchmarks, we report the results for the test set using the original answers as ground truth.

\begin{table}[htp]
\caption{Size and number of QA sets for the different splits of the Pirá 2.0 dataset. ``\#QAs'' stands for the number of QA sets.}
\label{tab:dataset_division}
\vspace*{3ex}
\centering
\setlength{\tabcolsep}{8pt}
\begin{tabular}{l|l|l}
\hline
{\bf Split} & \textbf{Size} & \textbf{\#QAs} \\ \hline 
Training & 80\% & 1806  \\
Development & 10\% &225  \\
Test & 10\% &227 \\ \hline
Full dataset & 100\% & 2258\\ 
\hline
\end{tabular}
\end{table}

\section{Closed Generative Question Answering}\label{sec:no_context}
To measure the specialized nature of the knowledge involved in Pirá 2.0, we establish our first benchmark, Closed Generative Question Answering (CGQA). The benchmark is given in the Definition (\ref{def:cgqa}) below.

\begin{definition}[\textit{CGQA}] 
\label{def:cgqa}
 In the CGQA benchmark, a question $q$ is provided, and a model has to output an answer $a'$ to it. Answer $a'$ is then compared to the annotated answer $a$.
\end{definition}

\subsection{Benchmark Setup}
Given the benchmark's definition, only generative models were considered. For English, we tested T5 \cite{raffel2019exploring} in its base and large version. T5 is a text-to-text (encoder-decoder) transformer model that can solve several NLP tasks  such as question answering and summarization. 
T5 has been pre-trained with the Colossal Clean Crawled Corpus (C4), a corpus two orders of magnitude larger than the Wikipedia, and went through subsequent training on several NLP tasks. This training procedure allowed T5 to reach state-of-the-art results in multiple tasks simultaneously, including question answering ones.

For the Portuguese part of the dataset, we tested two version of T5 adapted to Portuguese: PTT5 \cite{carmo2020ptt5} and mT5 \cite{xue-etal-2021-mt5}. PTT5 is a T5 model pretrained with BrWac \cite{wagner-filho-etal-2018-brwac}, a web corpus for Brazilian Portuguese. A potential drawback of this model is that PTT5 is not fine-tuned on QA tasks, unlike the original T5.
The second model tested for Portuguese was mT5. This is a multilingual version of T5, pretrained with a Common Crawl-based dataset covering 101 languages, including Portuguese. 

We also report the results for two LLMs, GPT3-turbo (aka ChatGPT) \cite{GPT3} and the recently released GPT4 \cite{openai2023gpt4}. These LLMs represent the state-of-the-art in foundational models (at least, for those commercially available). They have both demonstrated a number of impressive abilities, particularly as regards to in-context learning \cite{bubeck2023sparks}.

As a CGQA benchmark, only questions were provided as input, without the supporting texts. T5 (base and large), PTT5, and mT5 were fine-tuned for 40 epochs on Pirá 2.0' training set, using early stopping on the development set's loss. In every case, we used a learning rate of 2e-5 and gradient accumulation of 4 steps; batch size was set to the maximum available size that could be achieved for each model. Since questions and answers were relatively short, we fixed a Maximum Sequence Length (MLS) of 128 to both inputs and outputs for all fine-tuned models. GPT3-turbo and GPT4 were tested through zero-shot learning; i.e., when no examples are provided in the prompt. The templates used for the prompts are found in Appendix \ref{sec:appendix}. In all cases, the original questions were used as input and the original answers served as the ground truth (see the \textit{QA Creation} step in Section \ref{subsection:Pira2}).

\subsection{Results}
The performance in the CGQA benchmark was measured through the F1-score (F1) metrics.
F1-score compares the number of matched words between the annotated and predicted answers and returns a number between 0 and 1. To make the result more intuitive, the score is then re-scaled to the [0, 100] range. Differently from the analysis in SQuAD \cite{rajpurkar2016squad}, we did not exclude definite (``the'') and indefinite (``a'' and ``an'') articles from the answers before calculating the score. This is because automated methods of excluding articles from Portuguese texts can also remove critical words like prepositions and pronouns, which can impact the meaning of the answers.

Table \ref{tab:gpt_nocontext} displays the results for the CGQA benchmark.

\begin{table}[htp]
\caption{Performance in the CGQA benchmark, where no context is provided along the question. Input and output Maximum Sequence Length is 128 tokens. ``Language'' refers to the language of the question. EN = English; PT = Portuguese. Best result for each language is in bold.}
\label{tab:gpt_nocontext}
\vspace*{3ex}
\centering
\begin{tabular}{l|l|l}
\hline
\textbf{Model} & \textbf{Language} & \textbf{F1} \\ \hline
T5 base & EN & 12.91 \\
T5 large & EN & 11.96 \\
\textbf{GPT3-turbo} & \textbf{EN} & \textbf{15.48} \\
GPT4 & EN & 8.48 \\  \hline
\textbf{PTT5 base} & \textbf{PT} & \textbf{16.58} \\
mT5 & PT & 3.95 \\
GPT3-turbo & PT & 14.38 \\
GPT4 & PT & 8.27 \\ \hline
\end{tabular}
\end{table}

In English, the two T5 models performed similarly, with the base version scoring slightly better than the large version. GPT3-turbo achieved the highest F1-score in this language, while GPT4 obtained the worst results. In Portuguese, PTT5 achieved the best result, and again GPT3-turbo performed better than GPT4. The only multilingual model, mt5, obtained the lowest F1-score among all models.

To understand the results achieved by GPT3-turbo and GPT4, we conducted a qualitative analysis of the generated responses. Our analysis revealed that both models frequently avoided answering certain questions, either by stating that they were unable to answer it or that there was insufficient information to provide a response. Common responses included {\tt ``Unknown''}, {\tt ``Unknown without further information''}, and {\tt ``Not enough information provided''}. At least half of the questions did not receive a positive answer from either model. This contributed to their lower F1-scores, as they did not even attempt to answer some of the questions. 

Overall, no model achieved a high F1-score in the CGQA benchmark, demonstrating the importance of relying on a context for answering the quite technical questions from Pirá 2.0.

\section{Machine Reading Comprehension} \label{sec:mrc}

Pirá 2.0 is a SQuAD-like \cite{rajpurkar2016squad} dataset in which questions are accompanied by a supporting text that contains the answer.
For this reason, the second benchmark we created is related to Machine Reading Comprehension (MRC), following Definition~\ref{def:mrc_task}. 

\begin{definition}[\textit{MRC}] 
\label{def:mrc_task}
 In the MRC benchmark, both a question $q$ and a supporting text $t$ are provided, and a model has to output an answer $a'$ to question $q$ (usually a short contiguous text span from $t$). Answer $a'$ is then compared to the annotated answer $a$.
\end{definition}

Our approach to MRC is as follows. We begin by reporting human baselines for both English and Portuguese (Section \ref{sec:mrc_human}). For the other baselines in the MRC benchmark, our procedure differs for the two languages (Section \ref{sec:mrc_model}). We first produce baselines for the English part of the dataset, comparing the performance of extractive and generative models. Following that, we take the best performing model in English and apply it to the Portuguese part of the dataset. This approach, validated on a subset of tests, allows us to reduce the number of tests by half.
Preference is given to English because supporting texts are in this language, and  because a larger number of models  are available for it. To evaluate the performance of the MRC benchmark in Portuguese, supporting texts are automatically translated with the use of the Google Translator API. 

\subsection{Human Baselines} \label{sec:mrc_human}
To establish a reasonable upper limit in this benchmark, we compared the original and validation answers from Pirá 2.0 and used the former as ground truth. Table \ref{tab:human_baselines} shows the human baselines for Pirá 2.0. We also provide human baselines for three other well-known MRC datasets: SQuAD 1.1, SQuAD 2.0, and Natural Questions~\cite{kwiatkowski2019natural}, for the sake of comparison.

Table \ref{tab:human_baselines} reveals that the F1-scores for both English and Portuguese in Pirá 2.0 hover around the low 50s. While this constitutes a significant improvement over the results obtained in the CGQA task (Section \ref{sec:no_context}), where supporting text was not provided, the numbers for Pirá 2.0 are still considerably lower than those achieved by popular datasets, indicating that Pirá 2.0 captures a highly challenging task. For example, the human baselines for SQuAD 1.1 \cite{rajpurkar2016squad} and SQuAD 2.0 \cite{rajpurkar2018know} are 86.8 and 89.5, respectively, while Natural Questions has a lower bound of 73.4 for single annotators (and an upper bound of 87.2 for an ensemble of human annotators) \cite{kwiatkowski2019natural}. In general, the human baselines suggest that the MRC benchmark on Pirá 2.0 poses a substantial challenge for QA models.

\begin{table}[htp]
\caption{Human baselines for Pirá 2.0 and similar datasets in the MRC benchmark. ``Avg. \#words'' stands for the average number of words per answer. Values for SQuAD 2.0 and Natural Questions are extracted from \cite{nq_data} and  \cite{Sen_2020}, respectively; no value was found for SQuAD 1.1. For Pirá 2.0, the average number of words per answer is based on the test set; we use the original answer in the dataset and report values for both English and Portuguese.}
\label{tab:human_baselines}
\vspace*{3ex}
\centering
\begin{tabular}{l|l|l|c|l}
\hline
\textbf{Dataset} & \textbf{Language} & \textbf{Size} & \textbf{Avg. \#words} & \textbf{F1}\\ \hline
\multirow{2}{4em}{Pirá 2.0} & English    & \multirow{2}{4em}{2258} & 13.19  &  54.85  \\ 
& Portuguese & & 14.40 & 51.71 \\ 
 SQuAD 1.1 & English & 100K & - & 86.8 \\
 SQuAD 2.0 & English & 150K & 3.2 & 89.5  \\ 
  Natural Questions & English & 370K & 4 & 73.4  \\ \hline

\end{tabular}
\end{table}

We raise a number of hypotheses that could explain the low human baselines for Pirá 2.0 as compared to the other datasets:

\begin{itemize}
    \item The technical character of its supporting texts, which approach scientific subjects;
    \item The fact that answers are not restricted to spans of texts, which adds another layer of complexity to the task; the statistics for span (extractive) and non-span (generative) questions in Pirá 2.0 are displayed in Table \ref{tab:human_baselines}; 
    \item The size of answers in Pirá, which can be quite long, as shown in Table \ref{tab:spans}. This tends to harm our metrics, since with more words in an answer, the lower the probability of slicing the correct span. Also, with longer answers, more variation is possible; e.g., short/long answer, including exemplifications;
    \item Typical translation issues (as also noted with respect to Pirá 1.0 \cite{Paschoal2021}). In particular, names and nouns received different translations throughout the dataset, a behavior that directly affects quality metrics.
\end{itemize}



\begin{table}[h]
\caption{Statistics for span and non-span answers in the English part of the dataset. `S' represents span and `N' represents non-span. The original questions are from the QA Creation phase of Pirá 1.0, and the validation answers from the Assessment phase (Section \ref{sec:dataset}). The total number of questions in the original and validation sets differ because not all questions were assessed.}
\label{tab:spans}
\vspace*{3ex}
\centering
\begin{tabular}{l|l|l|l}
\hline
 & \textbf{Train} & \textbf{Development} & \textbf{Test} \\ \hline
 \multirow{2}{8em}{Original answer} & S: 238 (13.17\%)  &  S: 31 (13.77\%) &  S: 27 (11.01\%) \\ 
& N: 1568 (86.82\%) & N: 194 (86.22\%) & N: 200 (88.10\%) \\ 
\hline
\multirow{2}{8em}{Validation answer} & S: 372 (21.18\%) & S: 46 (21.39\%) & S: 47 (21.75\%) \\ 
& N: 1384 (78.81\%) & N: 169 (78.60\%) & N: 169 (78.24\%) \\  \hline
\end{tabular}
\end{table}

\subsection{Benchmark Setup} \label{sec:mrc_model}


For the English part of Pirá 2.0, two extractive models were tested: BERT and RoBERTa. BERT \cite{devlin2018bert} is a bidirectional encoder based on the transformer architecture. It is pretrained on unlabeled data for two tasks: masked language modeling and next sentence prediction. 
RoBERTa \cite{liu2019roberta} is a BERT model trained on additional data and with several hyperparameter optimizations,
such as pretraining with longer sentences and dynamical masking; it also dispenses the next sentence prediction task. As a consequence of its special architecture, RoBERTa achieves state-of-the-art results on SQuAD 1.1 and 2.0, GLUE \cite{wang-etal-2018-glue} and RACE \cite{lai-etal-2017-race}. 
On the generative side, we evaluated the performance of T5, in both its base and large versions.


For Portuguese, we first translated the textual passages from English so that the questions and supporting texts could be in the same language. As our extractive model, we used BERTimbau base \cite{souza2020bertimbau}, a BERT model trained for Brazilian Portuguese. On the generative side, we tested PTT5 base and mT5 base.

Extractive models were evaluated with a maximum sequence length (MSL) of 512 tokens. 
We did not fine-tune these models, as answers in Pirá 2.0 are not marked with the start and ending positions in the text. More importantly, our dataset contains mostly non-span answers, something that would prevent training. Instead, we took checkpoints for these models fine-tuned on SQuAD 1.1 and 2.0.

Generative models were fine-tuned with Pirá 2.0. The hyperparameter search was always carried out with respect to the development set. For English, we considered three MSL values for T5 base (512, 1024, 1536) and one for T5 large (512). For Portuguese, MSL was fixed at 1536 for PTT5 base and 512 for mT5. For the extractive models, whether in English or Portuguese, we adopted the MLS permitted, 512 tokens. 
In all cases, MSL and model size were chosen as to make the most of our available computational resources.

Finally, we also tested GPT3-turbo and GPT4 in the MRC benchmark. Our intention with this was to understand what can be obtained from the LMs and the relative advantage of fine-tuning a smaller model on a specific dataset. The two models are tested in a zero-shot learning environment, where both the context and the question are provided. The multilingual nature of these models also allowed us to test directly a combination of context in English and question in Portuguese (a plausible scenario for applications in low-resource languages). The templates used for the prompts are found in Appendix \ref{sec:appendix}.

\subsection{Results} \label{sec:mrc_results}
Results for the English part of the dataset are shown in Table \ref{tab:results_mrc_eng}. Among the extractive models, RoBERTa large achieved the best result. This indicates that model size and hyperparameter optimization both contribute to a model's performance in the MRC benchmark.

The choice of MSL also influenced results. Supporting texts in Pirá 2.0 are relatively long, often containing several paragraphs; because of that, models with limited input windows end up missing information that is critical to answering questions. We also observe that T5 large performed well below T5 base for the same MSL value of 512, as measured by the F1-score. We conjecture that this is due to the high number of parameters of T5 large (770M) as compared to the relatively small amount of training data from Pirá 2.0. Finally, we note that T5 base outperformed the best extractive model, RoBERTa large, in every scenario. Nonetheless, it is worth considering that RoBERTa achieved a F1-score of 48.22, as compared to T5's best score of 51.27, without undergoing any fine-tuning for Pirá 2.0. 

\begin{table}[t]
\caption{Comparison among extractive and generative models for MRC on the English part of Pirá 2.0. ``MSL" stands for  ``Maximum Sequence Length". Results were obtained on the test set; in bold, the best model for each type of model (extractive and generative).}
\label{tab:results_mrc_eng}
\vspace*{3ex}
\centering
\begin{tabular}{l|l|l|l|l}
\hline
\textbf{Model type} & \textbf{Model} & \textbf{Fine-tuned} & \textbf{MSL} & \textbf{F1} \\ \hline
Extractive & BERT base & SQuAD 1 & 512 & 41.54\\
Extractive & BERT large & SQuAD 2 & 512 & 46.96\\
Extractive & RoBERTa base & SQuAD 2 & 512 & 47.65 \\
\textbf{Extractive} & \textbf{RoBERTa large} & \textbf{SQuAD 2} & \textbf{512} & \textbf{48.22}  \\
\hline
Generative & T5 base & SQuAD 2/Pirá 2 & 512 & 49.12  \\
Generative & T5 base & SQuAD 2/Pirá 2 & 1024 & 50.50 \\
\textbf{Generative} & \textbf{T5 base} & \textbf{SQuAD 2/Pirá 2} & \textbf{1536} & \textbf{51.27} \\
Generative & T5 large & SQuAD 2/Pirá 2 & 512 & 41.22 \\ \hline
\end{tabular}
\end{table}

Results for the Portuguese part of the dataset are shown in Table \ref{tab:results_mrc_pt}. The F1-score for PTT5 was almost two times larger than that of mT5, showing that this multilingual model is not yet prepared for complex MRC tasks in Portuguese. PTT5 falls short too in the comparison with models trained for English: its F1-score is around half of that of the T5 base models. This lower score is largely due to its lack of fine-tuning on QA tasks.


\begin{table}[t]
\caption{Comparison among extractive and generative models for MRC on the Portuguese part of Pirá 2.0. ``MSL" stands for  ``Maximum Sequence Length". Results were obtained on the test set; in bold, the best model for each type of model (extractive and generative).}
\label{tab:results_mrc_pt}
\vspace*{3ex}
\centering
\begin{tabular}{l|l|l|l|l}
\hline
\textbf{Model type} & \textbf{Model} & \textbf{Fine-tuned} & \textbf{MSL} & \textbf{F1}\\ \hline
\textbf{Extractive} & \textbf{BERTimbau} & \textbf{Squad 1.1} & \textbf{512} & \textbf{37.53}\\ \hline
\textbf{Generative} & \textbf{PTT5 base} & \textbf{Pirá 2} & \textbf{1536} & \textbf{27.90}  \\ 
Generative & mT5 base & SQuAD 2/Pirá 2 & 512 & 14.23 \\ \hline
\end{tabular}
\end{table}

Finally, in Table \ref{tab:mcr_gpt}, we display the results for GPT3-turbo and GPT4 in the MCR benchmark . The benchmark was conducted under three different scenarios: the first two scenarios involved both the supporting text and the question being in the same language, while in the last scenario, the supporting text was in English and the question was in Portuguese. This scenario simulates a common real-life situation where the database is in a different language than that spoken by the users. The objective of this test was to determine whether LLMs require prior translations or if they are capable of directly handling information in multiple languages.

GPT-4 exhibited superior performance in English-to-English scenarios, although its effectiveness decreased when Portuguese was introduced. However, even in English-to-Portuguese cases, GPT-4 delivered results that were only marginally lower than those achieved in Portuguese-to-Portuguese scenarios. Notably, both GPT-3 Turbo and GPT-4 demonstrated the ability to comprehend information in multiple languages, suggesting that Language Models (LLMs) could potentially eliminate the need for a translation step. Nevertheless, when compared to fine-tuned models, both GPT-3 Turbo and GPT-4 achieved relatively lower scores. This was primarily due to their tendency, similar to the behavior observed in the CGQA benchmark (Section \ref{sec:no_context}), to withhold responses to certain questions, often providing answers like {\tt ``Unknown''} or {\tt ``Not enough information provided''}. Such behavior significantly impacted their F1-scores, thereby explaining their comparatively poorer performance.

\begin{table}[h]
\caption{Performance of GPT3-turbo and GPT4 in the MCR benchmark. ``Text'' refers to the language of the supporting language and ``Question'' refers to the language of the question. EN = English; PT = Portuguese. In bold, the best result for each language combination.}
\label{tab:mcr_gpt}
\vspace*{3ex}
\centering
\begin{tabular}{l|l|l|l}
\hline
\textbf{Model} & \textbf{Text} & \textbf{Question} & \textbf{F1}\\ \hline
GPT3-turbo & EN & EN & 5.90 \\
\textbf{GPT4} & \textbf{EN} & \textbf{EN} & \textbf{7.90}\\ \hline
\textbf{GPT3-turbo} & \textbf{PT} & \textbf{PT} & \textbf{8.78}\\
GPT4 & PT & PT & 7.62\\ \hline
\textbf{GPT3-turbo} & \textbf{EN} & \textbf{PT} & \textbf{8.43} \\ 
GPT4 & EN & PT & 7.40\\ \hline
\end{tabular}
\end{table}

\section{Information Retrieval} \label{sec:ir}

The third benchmark we have studied is one related to Information Retrieval (IR), as given by Definition~\ref{def:ir}:

\begin{definition}[\textit{IR}] 
\label{def:ir}
In the IR benchmark, a model needs to traverse a corpus $C$ and deliver the $k$ most relevant supporting texts for answering question $q$, where $S = \{s_1, ..., s_k\}$ is the set of retrieved supporting texts. One then checks whether the supporting text $t_q$ that was used to create question $q$ is in $S$.
\end{definition}


As noted previously, questions in Pirá 2.0 are tied to a supporting text (a scientific abstract or small excerpt of text). In order to examine this dataset in the context of a traditional IR task, we gather all supporting texts in a corpus $C$. Given a question $q$, an IR model then must search through corpus $C$ and retrieve the $k$ most relevant supporting texts that should contain the answer $a$, where $S = \{s_1, ..., s_k\}$ is the set of retrieved texts. The evaluation process consists of checking whether the supported text for $q$, $t_q$, is among the retrieved texts in $S$.
The integer $k$ is a hyperparameter of the model.

\subsection{Benchmark Setup}

Two types of IR models were investigated for this benchmark: BM25 \cite{Robertson2009}, a sparse retrieval, and Dense Passage Retrieval (DPR) \cite{karpukhin-etal-2020-dense}, a dense one.
BM25 treats sentences as bag-of-words, similarly to Term Frequency–Inverse Document Frequency (TF-IDF), but giving more weight to longer texts. BM25 is a fast algorithm that does not require any training. A disadvantage of sparse methods like this, however, is that they are not able to consider semantic information when retrieving texts. Dense methods such as DPR, instead, rely on converting texts and queries to embeddings through a language model (e.g., BERT), and measuring their similarity. This allows for semantic-based text retrieval, a capability that sparse methods like BM25 lack.

\subsection{Results}
\begin{figure}[htp]
  \centering
  \includegraphics[width=0.65\textwidth]{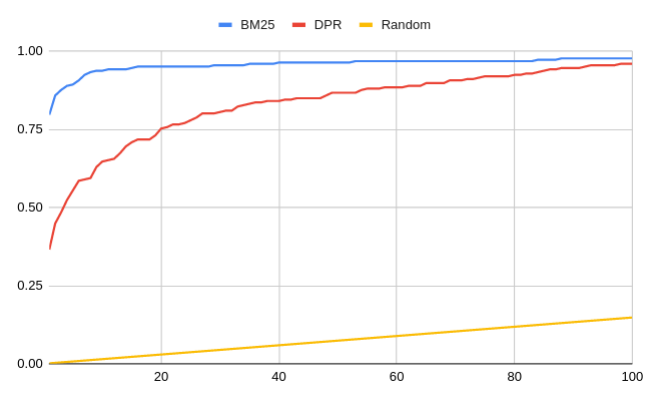}
  \caption{Performance comparison between BM25 (sparse, in blue) and DPR (dense, in red) for increasing number of retrieved texts $k$. In yellow, we plot the performance of a simulated random retriever. The x-axis give the accuracy of the retrieval methods and the y-axis the different values of k. }
  \label{fig:bm25_v_dpr}
\end{figure}

We first compare the performance of BM25 and DPR in the English version of the dataset. For each question $q$, a model has to retrieve $k$ supporting texts, and if one of them contains the answer to the question, we have a success. Figure \ref{fig:bm25_v_dpr} shows the performance of each retrieval method for $k$ ranging from 1 to 100. The figure also plots the performance of a simulated random retrieval, in which the value of the y-axis is $k/C$ (yellow line), representing the minimum value a model should get (considering that it can retrieve the correct passage by chance). As can be seen, BM25 consistently surpasses DPR for any value of $k$. Furthermore, BM25 reaches more than 90\% of accuracy for $k \geq 6$.
Note that  our experiment  used  DPR's standard version, pre-trained on Wikipedia, without fine-tuning on our domain corpus.

BM25 was expected to outperform DPR in Pirá 2.0, as it had been shown to achieve better results in SQuAD 1.1 \cite{karpukhin-etal-2020-dense}, a dataset that is similar to Pirá in a number of ways. The   authors of DPR suggest  that neural retrieval models have superior performance in cases where the question and the correct text do not necessarily share the same words, since DPR can grasp the context and meaning of texts, while keyword methods simply search by matching words. Nonetheless, annotators of Pirá 2.0 often used snippets from the supporting texts to create answers. As a consequence, it was common for texts and answers to share words --- a pattern that favored BM25, a sparse model.

As DPR is not currently available in Portuguese, our testing for the IR benchmark was restricted to variations of BM25. However, since supporting texts were in English, we had to make some adaptations. Two different approaches were tested. In the first, we used the Google Translator API to translate the Portuguese questions into English and then queried the original corpus. In the second, we translated the supporting texts into Portuguese (as described in Section \ref{sec:mrc}), and used the original questions in Portuguese to query the translated corpus.

\begin{figure}[htp]
  \centering
  \includegraphics[width=0.65\textwidth]{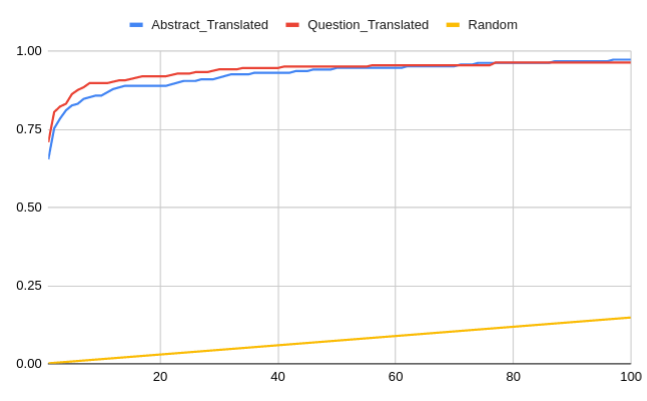}
  \caption{Performance comparison between the two different implementations of BM25 for Portuguese.  In the first, the system searches for supporting texts that contain the correct answer in a Portuguese-translated version of the corpus (blue line). In the second, the system translates the question into English and queries the original English corpus (red line). In yellow, we plot the performance of a simulated random retriever. The x-axis give the accuracy of the retrieval methods and the y-axis the different values of k.}
  \label{fig:translated_paragraphs}
\end{figure}

In Figure \ref{fig:translated_paragraphs}, we compare the two strategies used in our experiments: translating the entire corpus into Portuguese (represented by the blue line) and translating only the questions into English (represented by the red line). The yellow line represents a simulated random retrieval to give an idea of the relative success of the retrievers. Overall, there is little difference between the two methods, with the translation of the questions being slightly more effective than the translation of the entire corpus for $k < 50$. This discrepancy could be attributed to the larger size of the supporting texts, resulting in more variability in the translations.

\section{Open Question Answering} \label{sec:oqa}

After exploring MRC and IR benchmarks, it becomes natural to create an Open Question Answering (OQA) benchmark that combines these two tasks. In OQA, a model is presented with a question as input and is allowed to query an external data source, such as a large corpus, to retrieve relevant supporting texts. The selected texts are then used to generate an answer to the question. Hence, OQA can be viewed as a combination of the previous two benchmarks: an IR task for finding relevant texts and an MRC task for converting the selected texts into a final answer. Definition~\ref{def:OQA} formally states the OQA benchmark.
 
\begin{definition}[\textit{OQA}] 
\label{def:OQA}
 The OQA benchmark comprises two steps. First, given a question $q$, a model has to traverse a corpus $C$ and deliver the $k$ most relevant passages $p$ from corpus $C$ for answering question $q$, where $P = \{p_1, ..., p_k\}$ is the set of retrieved passages. Next, given question $q$ and the $k$ most relevant passages $P$, a model has to output an answer $a'$ to question $q$. Answer $a'$ is then compared to the annotated answer $a$.
 \end{definition}

Unlike the MRC benchmark, the IR benchmark divides the corpus C into passages rather than using the original supporting texts. This is because reader models often have a limited input capacity and may struggle to process multiple supporting texts in full. To address these issues, we adopted the approach described in \cite{caccao2021deepage, karpukhin-etal-2020-dense}, using Haystack, an open-source NLP framework, to segment the corpus into short passages of 100 words.\footnote{\url{https://haystack.deepset.ai/}.}

\subsection{Benchmark Setup}
Our baseline solution is built upon the DEEPAGÉ \cite{caccao2021deepage} architecture, which combines an IR module based on BM25 and a language generation model --- T5 or PTT5, depending on the language. For the English part of Pirá 2.0, we adopted the top-performing IR method identified in Section \ref{sec:ir}, BM25, and combined it with the two best-performing readers from the MRC benchmark, namely RoBERTa (extractive) and T5 (generative), as detailed in Section \ref{sec:mrc}. Similarly, for the Portuguese part, we used BM25 as the retriever and chose one model of each type: BERTimbau (extractive) and PTT5 (generative). Notably, we did not test GPT3-turbo and GPT4 in this benchmark, as they lack retrieval capabilities.


\subsection{Results}

While the OQA benchmark is more challenging than the MRC benchmark by definition, as it involves an additional retrieval component, we were able to achieve reasonable results in OQA by combining the best reader and retriever models found in the previous benchmarks. In fact, the F1-scores for OQA in English are only slightly worse than those obtained in MRC. Table \ref{tab:results_oqa_eng} shows the F1-scores for the English OQA benchmark. For RoBERTa large, the F1-score went from 48.22 in MRC to 41.65 in OQA (a decrease of 13.62\% for $k = 5$). Similarly, the F1-score for the T5 base model went from 51.27 in MRC to 48.11 in OQA (a decrease of 6.16\% for $k = 15$).

In the OQA benchmarks, generative models performed much better than extractive ones. The difference between the best generative and extractive models in this task was $15.51\%$, compared to a difference of just $6.33\%$ in MRC. This behavior is possibly due to the greater flexibility to generate responses from $k$ passages than from a single text. Moreover, for generative models, increasing $k$ positively impacts the quality of answers. For the three different values of $k$ we tested (5, 10 and 15), T5 had its best performance with $k = 15$. In the case of RoBERTa,  there is almost no difference between the different values of $k$. This is due to the limited value of MSL adopted: the model's ability to process texts was saturated when reaching 512 input tokens, and everything else was ignored from there on. In T5, on the contrary, we were able to increase MSL up to 1536, allowing a larger amount of passages to be processed.

\begin{table}[htp]
\caption{Comparison of extractive and generative models for OQA on the English part of Pirá 2.0. We adopted 512 tokens as the Maximum Sequence Length for extractive models and 1536 tokens for generative ones. The number of retrieved passages is indicated by $k$. Results were obtained with the test set; in bold, the best model for each type (extractive or generative).}
\label{tab:results_oqa_eng}
\vspace*{3ex}
\centering
\begin{tabular}{l|l|l|l|l}
\hline
\textbf{Model type} & \textbf{Model} & \textbf{Fine-tuned} & \textbf{k} & \textbf{F1} \\ \hline
\textbf{Extractive} & \textbf{RoBERTa Large} & \textbf{SQuAD 2} & \textbf{5} & \textbf{41.65} \\
Extractive & RoBERTa Large & SQuAD 2 & 10 & 40.87  \\
Extractive & RoBERTa Large & SQuAD 2 & 15 & 40.48 \\ \hline
Generative & T5 Base & SQuAD 2/Pirá 2 & 5 & 45.99 \\
Generative & T5 Base & SQuAD 2/Pirá 2 & 10 & 47.12  \\
\textbf{Generative} & \textbf{T5 Base} & \textbf{SQuAD 2/Pirá 2} & \textbf{15} & \textbf{48.11} \\ \hline
\end{tabular}
\end{table}

To address the Portuguese part of Pirá 2.0, we chose to translate the abstract to Portuguese, as explained in Section \ref{sec:oqa}, and then fed the resulting concatenated string to reader models in the same language.\footnote{The other strategy, i.e., translating the question into English and then querying the texts, demands an additional step: finding the corresponding passages of the retrieved texts in the Portuguese-translated corpus before passing them to the reader models. } As our readers, we evaluated BERTimbau and PTT5, an extractive and generative reader, respectively. For the latter, we used the same hyperparameters as for the best performing T5 model in English, with $k$ set to 15 passages.

Results for this experiment are presented in Table \ref{tab:results_oqa_pt}. 
Contrary to what one would expect, the F1-score in the OQA benchmark for BERTimbau and PTT5 base were higher than those in the MRC benchmark. This is because questions in Pirá 2.0 often do not require the entire supporting text to produce a correct answer. Consequently, the model can overlook relevant information when dealing with lengthy texts. By dividing the corpus into passages, we can increase the likelihood of finding the necessary information to answer a question, particularly in the case of factual questions.

\begin{table}[htp]
\caption{Comparison of extractive and generative models for OQA on the Portuguese part of Pirá 2.0. The Maximum Sequence Length was set to 1536 tokens for PTT5 and 512 for BERTimbau. The number of retrieved passages is indicated by $k$.}
\label{tab:results_oqa_pt}
\vspace*{3ex}
\centering
\begin{tabular}{l|l|l|l|l}
\hline
\textbf{Model type} & \textbf{Model} & \textbf{Fine-tuned} & \textbf{k} & \textbf{F1}\\ \hline
\textbf{Extractive} & \textbf{BERTimbau} & \textbf{Squad 1.1} & \textbf{15} & \textbf{31.23}\\ \hline
\textbf{Generative} & \textbf{PTT5 Base} & \textbf{Pirá 2} & \textbf{15} & \textbf{24.47}\\ \hline
\end{tabular}
\end{table}

\section{Answer Triggering} \label{sec:AT}

In the MRC and QA benchmarks, every input question receives an answer from the system in the form of a non-empty string.
In practice, however, no QA system is capable of answering all questions provided by users (except, of course, if questions are picked from a closed set). Questions may be ill-formed or ambiguous; the databases may not have the necessary knowledge; or  contextual information may be implicitly assumed by speakers. 
In those cases, the policy of always answering questions may be detrimental. Indeed, when stakes are high, a QA system should refrain from answering questions unless strongly certain of the response.\footnote{There is a large literature on the harmful impact that language models can have on bias amplification and toxicity \cite{weidinger2021ethical}. In this paper, we consider the more limited problem of inaccurate answers.}
A more cautious policy is important to guarantee alignment and factual grounding~\cite{thoppilan2022lamda}. 

Within this context,
the fifth benchmark we established was Answer Triggering (AT). This is a binary classification task where one has to decide whether a question can be answered or not, given the available information. AT has received growing attention since the release of SQuAD 2.0, a dataset that added non-answerable questions to the original SQuAD 1.1 dataset \cite{rajpurkar2018know}. The formal definition of the AT benchmark is stated in Definition~\ref{def:AT} below:

\begin{definition}[\textit{AT}] 
\label{def:AT}
In the AT benchmark, a question $q$ and a supporting text $t$ are provided, and a model has to output a label $l'$, indicating whether the question $q$ can be answered (1) or not (0), $l \in \{0,1\}$. Label $l'$ is then compared to the annotated label $l$. 
 \end{definition}

By allowing some questions to remain unanswered, models are required to learn not only \textit{how} to answer a given question but also \textit{when} to do so. This filtering mechanism ensures a better control of the answers given by a QA system and the trust placed on it.\footnote{GPT3-turbo and GPT4 tend to avoid answering some questions due to their training, as described in Sections \ref{sec:no_context} and \ref{sec:mrc}. However, unlike the AT benchmark, this behavior is not intentionally programmed or controlled.}


\subsection{Benchmark Construction}
In standard AT datasets, unanswerable questions are intentionally constructed, as is the case with SQuAD 2.0 \cite{rajpurkar2018know}, WikiQA \cite{yang2015wikiqa}, and SelQA~\cite{jurczyk2016selqa}. Pirá 2.0 does not contain intentionally produced unanswarable questions; by construction, questions are paired with their (putative) answers. 
Pirá 2.0 also does not include indications of uncertainty, as found in datasets such as ReCO \cite{wang2020reco} and QuAIL \cite{rogers2020getting}. Therefore, in order to generate an AT version of Pirá 2.0, we relied on the information provided in the Assessment phase to select unanswerable questions (see Secion \ref{subsection:Pira1}). In this stage, participants had to evaluate whether the QA sets created in the previous phase were meaningful, according to a Likert scale.\footnote{(1) Strongly disagree, (2) Disagree, (3) Neither agree nor disagree, (4) Agree, (5) Strongly agree.} For our purposes, we take question meaninglessness as a proxy for unanswerability: we consider that meaningless questions cannot be properly answered. The AT benchmark is thus defined as a binary classification task where 1 means that a question has an answer and 0 that it has not. For the labeling process, QA sets with an assessment of 4-5 are classified as answerable and those with an assessment of 1-2 as unanswerable; QA sets with an assessment of 3 are discarded, given their neutrality.\footnote{A subset of the present authors has studied AT in Pirá 2.0 from the viewpoint of regression, where meaningfulness is treated as a continuous scale. They show that filtering unfit questions contribute to the quality of the answers generated by an answerer system \cite{pirozelli2022AT}.} 


The AT benchmark suffers with the small number of unanswerable questions from Pirá 2.0. For this reason, we employed two data augmentation techniques as an attempt to remediate the lack of data. In the first approach, the training set was enlarged with the validation answers and the paraphrases of questions produced in the assessment phase of Pirá 1.0 and revised for Pirá 2.0. They were then combined with the original questions and answers to generate the maximum possible number of permutations in the training set. In the second approach, the training set was augmented through the incorporation of automatically-generated paraphrases. Two paraphrases were produced for each (original) question, which were again combined with manual questions and answers (original and paraphrased) in all possible ways. Automatic paraphrases in English were generated with PEGASUS \cite{pegasus2020} and those in Portuguese with PTT5-Paraphraser \cite{pellicer2022paraphraser}. The number of QA sets for each version of the dataset is presented in Table \ref{tab:AT_datasets}.

\begin{table}[htp]
\caption{Number of QA sets divided by label for the different types of dataset construction. The \textit{Standard} dataset is the original Pirá 2.0 dataset after the answer triggering labeling process.  In the \textit{Hum. Paraphrase} dataset, the training set is expanded with the paraphrases for questions generated in the Assessment phase of Pirá 1.0. In the \textit{Autom. Paraphrase} dataset, the training set is augmented with automatic paraphrases for questions, using PEGASUS for English and PTT5-Paraphraser for Portuguese. ``A'' denotes answerable questions, ``U'' denotes unanswerable questions, and ``T'' represents the total number of questions. The total number of questions for AT is lower than the original splits (Table \ref{tab:dataset_division}), as some QA sets were not assessed for question meaningfulness and neutral questions (scored 3 on a 1 to 5 scale) were removed from the labeling process.}
\label{tab:AT_datasets}
\centering
\begin{tabular}{l|lll|lll|lll}
\hline
\multicolumn{1}{l|}{\multirow{2}{*}{\textbf{Split}}} & \multicolumn{3}{c|}{\textbf{Standard}} & \multicolumn{3}{c|}{\textbf{Hum. Paraphrase}}   & \multicolumn{3}{c}{\textbf{Autom. Paraphrase}}\\
& \textbf{A} & \textbf{U} & \textbf{T} & \textbf{A} & \textbf{U} & \textbf{T} & \textbf{A} & \textbf{U} & \textbf{T}\\
\hline
Training & 1432 & 145 & 1577 & 2733 & 265 & 2998 & 5597 & 555 & 6152\\
Development & 173 & 19 & 192 & 173 & 19 & 192 & 173 & 19 & 192 \\
Test & 179 & 19 & 198 & 179 & 19 & 198 & 179 & 19 & 198 \\
\hline
\end{tabular}
\end{table}

\subsection{Benchmark Setup}

Four sets of baselines were considered in this benchmark. In every case, a concatenation of supporting text and question was used as the input. The first baseline assigned all observations in the test set to the largest class; in this case, answerable questions. Note that QA sets strongly leaned to meaningfulness according to the human assessments of Pirá 2.0. Hence, to perform better than that, a model should capture subtler elements that indicated unanswerability.

Following Jos\'e et al.\ \cite{jose2022}, we also tested a Naïve Bayes classifier. Such technique is useful in text classification and sentiment analysis, as classes in those tasks can be often determined by the occurrence of certain words. We used Naïve Bayes as a baseline for measuring how much AT is dependent on vocabulary; if this algorithm performs well, there is evidence that unanswerability can be detected by the presence of specific words.

Next, we tested a number of transformer-based models, which can take word order and context into account. For English, BERT and RoBERTa were used, in their base and large versions. For Portuguese, we used BERTimbau \cite{souza2020bertimbau} base. Models were fine-tuned for 8 epochs on Pirá 2.0's training set, using early stopping on the development set's loss. In every case, we used a learning rate of 2e-5 and gradient accumulation of 4 steps; batch size was set to the maximum available size that could be achieved for each model.

Finally, we tested GPT3-turbo and GPT4 in the AT benchmark through zero-shot learning. Given a question and a context, the models were instructed to output 0 if the question could not be answered with the provided context, and 1 if it could be answered. The templates used for the prompts are found in Appendix \ref{sec:appendix}.

\subsection{Results}

To assess our results, we utilize the F1-score in its standard definition for classification tasks; i.e., the harmonic mean of precision and recall. Unlike accuracy, which may be biased towards the majority class in imbalanced datasets, the F1-score is suitable for evaluating performance when there is a class imbalance.
Table \ref{tab:AT_baselines_geral} presents the results for the class assignment and Naïve Bayes baselines. The majority class assignment achieved an F1-score of 85.84. The Naïve Bayes classifier, like the class assignment baseline, assigned all instances to the majority class. The dataset augmentations did not improve the performance of this classifier. These findings suggest that answerability is not trivially related to the presence of specific words.

\begin{table}[htp]
\caption{F1-score for class assignment and Naïve Bayes in the AT benchmark. The \textit{Standard} dataset is the original Pirá 2.0 dataset after the answer triggering labeling process.  In the \textit{Hum. Paraphrase} dataset, the training is expanded with the paraphrases for questions generated in the Assessment phase of Pirá 1.0. In the \textit{Autom. Paraphrase} dataset, the training set is augmented with automatic paraphrases for questions, using PEGASUS for English and PTT5-Paraphraser for Portuguese.}
\label{tab:AT_baselines_geral}
\centering
\begin{tabular}{l|c|c}
\hline
\textbf{Dataset} & \textbf{Class assignment} & \textbf{Naïve Bayes}\\ \hline
Standard & \multirow{3}{*}{\textbf{85.84}} & \multirow{3}{*}{\textbf{85.84}} \\
Hum. Paraphrase   &  &  \\
Autom. Paraphrase &  &  \\
\hline
\end{tabular}
\end{table}

Table \ref{tab:AT_baselines_transformer} presents the results obtained with the transformer-based classifiers. The F1-score ranged from 83.43 to 85.85 for the English part of Pirá 2.0 and from 84.31 to 85.59 for Portuguese. As with the previous baselines, all models assigned the majority of observations to the answerable class, with minor exceptions accounting for the slight variations in their F1-scores. Like the other baselines, transformer models were unable to effectively solve the AT benchmark. Furthermore, the addition of manual or automatic paraphrases did not provide any clear improvements.


\begin{table}[htp]
\caption{F1-score for transformer models in the AT benchmark. The \textit{Standard} dataset is the original Pirá 2.0 dataset after the answer triggering labeling process.  In the \textit{Hum. Paraphrase} dataset, the training set is expanded with the paraphrases for questions generated in the Assessment phase of Pirá 1.0. In the \textit{Autom. Paraphrase} dataset, the training set is augmented with automatic paraphrases for questions, using PEGASUS for English and PTT5-Paraphraser for Portuguese. BERT anr RoBERTa are used in the English part of the dataset, whereas BERTimbau is used in the Portuguese part. Best results for each language in bold.}
\label{tab:AT_baselines_transformer}
\centering
\begin{tabular}{l|ll|ll|c}
\hline
\multicolumn{1}{l|}{\multirow{2}{*}{\textbf{Dataset}}} & 
\multicolumn{2}{c|}{\textbf{BERT}} & 
\multicolumn{2}{c|}{\textbf{RoBERTa}} & 
\multicolumn{1}{l}{\multirow{2}{*}{\textbf{BERTimbau}}}\\   
& \textbf{Base} & \textbf{Large} & \textbf{Base} & \textbf{Large} &  \\
\hline
Standard & \textbf{85.85} & 84.31 & 85.84 & 85.84 & 84.31\\
Hum. Paraphrase & 83.43 & 85.84 & 84.98 & 85.56 & \textbf{85.59}\\
Autom. Paraphrase & 83.84 & 85.84 & 85.56 & 85.84 & 85.08\\
\hline
\end{tabular}   
\end{table}

Results for GPT3-turbo and GPT4 are given in Table \ref{tab:GPT_AT}. These models were able to distinguish between answerable and unanswerable questions to some extent. GPT3-turbo achieved an F1-score of 87.37 for English and 89.39 for Portuguese, which represents a 1.77\% and 4.12\% improvement, respectively, over the majority class baseline. Although these models still assigned most questions to the answerable class, they correctly identified some questions as unanswerable.


Overall, further investigation is needed to assess the quality of the annotations in Pirá 2.0. Additionally, a larger dataset is needed to establish a more reliable benchmark for the models requiring fine-tuning.

\begin{table}[htp]
\caption{Performance for GPT3-turbo and GPT4 in the AT benchmark. ``Language'' represents the language of both the supporting text and the question. ``Label'' breaks the predicted labels by answerable and unanswerable questions, respectively.}
\label{tab:GPT_AT}
\centering
\begin{tabular}{l|c|c|c}
\hline
\textbf{Model} & \textbf{Language} & \textbf{F1} & \textbf{Label}\\
\hline
\textbf{GPT3-turbo} & \textbf{EN} & \textbf{87.37} & \textbf{190 / 8}\\
GPT4 & EN & 85.35 & 186 / 12\\
\hline
\textbf{GPT3-turbo} & \textbf{PT} & \textbf{89.39} & \textbf{192 / 6}\\
GPT4 & PT & 85.85 & 187 / 11\\
\hline
\end{tabular}
\end{table}


\section{Multiple-Choice Question Answering} \label{sec:mcqa}
The last benchmark we investigated is related to Multiple-Choice Question Answering (MCQA). In a standard QA task, the aim is that of finding (or generating) a single correct answer for a question (or, at most, several equivalent answers for each question, such as in SQuAD 1.1). In multiple-choice QA, on the contrary, several different answers are provided for each question, and the goal is to pick the correct one among the set of  alternatives. This benchmark is formally stated by Definition~\ref{def:MCQA} above:

\begin{definition}[\textit{MCQA}] 
\label{def:MCQA}
 In the MCQA benchmark, a question $q$, a supporting text $t$, and $k$ candidate answers $a_k \in Z_q$ are provided, where $Z_q$ is the set of candidate answers for question $q$. One and only one candidate answer $a_k \in Z_q$ is true. A model has to output a candidate answer $a_k'$. Answer $a_k'$ is then compared to the annotated answer $a_k$.
 \end{definition}

Our goal with this benchmark is to assess the ability of machine learning models to find the correct answer to a question when several plausible answers are provided.

\subsection{Benchmark Construction} \label{qa_mcqa}

MCQA datasets are not as easy to construct as MRC and OQA ones. A challenging MCQA dataset relies on plausible candidate alternatives which may distract from the correct responses; otherwise, the task becomes trivial. For this reason, producing a MCQA dataset from scratch demands complex and costly  procedures to generate alternatives. Due to this, most MCQA datasets are built around standardized educational tests, such as GRE and LSAT \cite{clark2016combining, clark2018think, lai2017race, MedMCQA2022}, or produced through crowdsourcing work \cite{richardson2013mctest, ostermann-etal-2018-semeval}.


With these issues in mind, we devised a method for creating a challenging MCQA dataset out of Pirá 2.0, taking inspiration from the long-standing tradition in AI of automatically generating alternatives for MCQA  \cite{EarlyMCQAgenerator, GeneratingMCQAsurvey}. What sets our approach apart from standard MCQA construction methods is that we adapt a MRC dataset to a MCQA format. Unlike standard methods, which need to generate not only candidate answers, but questions and correct answers as well, we focus on generating plausible (but ultimately wrong) candidate answers or distractors. Our methodology closely resembles that used in \cite{RecipeQA}, with the difference that they selected distractors randomly, whereas we opted for an IR approach that enables us to choose candidate answers that resemble the correct answer.

The set of candidate answers in our MCQA dataset consists of the correct answer to a question as well as other answers in the dataset that have a high overlap with the supporting text. This approach ensures that the candidate answers are well-formed sentences and lexically resemble the correct answer.
More formally, 
in order to create the MCQA instances, we index all the answers in the dataset, $a_i \in A$, where $A$ stands for the set of all answers in Pirá 2.0. Then, for each supporting text $t_j \in T$, where $T$ is the set of all supporting texts in Pirá 2.0, we establish the similarity of each answer $a_i$ to each text $t_j$. We use TF–IDF to represent the content of answers and supporting texts, and employ cosine similarity to measure their similarity.
Next, for each text $t_j$, we retrieve the $k$ answers in $A$ most similar to it, creating $Z_j$, the set of candidate answers to $t_j$; in our dataset, we fix $k = 5$ as the number of candidate answers per question. Finally, we check whether $a_j$, the correct answer to $t_j$, is in $Z_j$. If not, we replace the least similar answer $a_k$ in $Z_j$ by $a_j$, as to guarantee that the correct answer is among the set of candidate answers. Finally, the set of candidate answers is shuffled.\footnote{For the texts with multiple questions (and, therefore, multiple correct answers, one for each question), we generalize the procedure described above. For the $m$ questions for text $t_j$, $Q_j = \{q_{j1}, ..., q_{jm}\}$, we check whether answer $a_{jm}$ is in the set of candidate answers $Z_j$; if not, we remove the answer $a_n \in Z_j$ with the smallest overlap with text $t_j$ that is not an answer for a question in $Q_j$, and replace it by $a_{jm}$. Thus, all questions about the same text end up in the same candidate answer set, ensuring that the candidate answers are all related to the same subject. For texts with more than five questions, each group of five correct answers was organized as a single MCQA instance, and the remaining answers were gathered with candidates obtained through the IR method previously described, as to form a regular MCQA instance.}

Table \ref{tab:questions_per_text} presents statistics on the number of questions per text for the three different splits (training, development, and test), as well as for the entire dataset.

\begin{table}[htp]
\caption{Statistics for the number of questions in Pirá 2.0. ``Max. \#question'' indicates the maximum number of questions associated with a supporting text. ``Avg. \#question'' denotes the average number of questions associated with supporting texts.}
\label{tab:questions_per_text}
\vspace*{3ex}
\centering
\begin{tabular}{l|c|c}
\hline
\textbf{Split} & \textbf{Max. \#questions} & \textbf{Avg.  \#questions} \\ \hline
Training & 17 & 2.81
\\
Development & 3 & 1.24
\\
Test & 4 & 1.25
\\ \hline
Full dataset & 23 & 3.34
\\ \hline
\end{tabular}
\end{table}


Table \ref{tab:answer_Kvalues} displays the percentage of times that the correct answer appears among the first \textit{k} positions in the similarity ranking, prior to replacement. While correct answers often rank highly, their overlap with the supporting texts is not always perfect. By constructing the dataset in this manner, we ensure that achieving high performance necessitates a certain level of comprehension of the question, rather than mere word associations. Additionally, the fact that supporting texts in Pirá 2.0 typically contain several questions and are drawn from a narrow domain contributes to the creation of plausible candidate alternatives.



\begin{table}[htp]
\caption{Percentage of times that the correct answer appears among the first \textit{k} positions of the similarity ranking (prior to replacement), for each split and for the full dataset.}
\label{tab:answer_Kvalues}
\vspace*{3ex}
\centering
\begin{tabular}{l|c|c|c}
\hline
\textbf{Split} & \textbf{\textit{k} = 1} & \textbf{\textit{k} = 5} & 
\textbf{\textit{k} = 50} \\ \hline
Training & 27.58\% & 63.23\% & 
88.82\%\\
Development & 56.00\% & 82.66\% & 
94.66\%\\
Test & 57.70\% & 85.02\% & 
95.59\%\\ \hline
Full dataset & 33.34\% & 67.13\% & 
89.76\% \\
\hline
\end{tabular}
\end{table}

\subsection{Benchmark Setup}\label{solving_mcqa}
As an initial baseline, we consider a random guesser that has a 20\% chance of selecting the correct answer from the five candidate alternatives. This represents a minimum level of performance in the MCQA benchmark. 
In addition, we employ three other baseline methods: an IR method; a transformer-based approach, UnifiedQA \cite{unifiedqa}; and two LLMs, GPT3-turbo and GPT4. Since there are no available versions of UnifiedQA or similar models for Portuguese, the MCQA benchmark is currently limited to English.

Our IR approach is based on  \cite{ENEM-Challenge, clark2016combining}.
For each question in the dataset,
we concatenate the question with each of the five candidate answers and compare their similarity to the supporting text. Next, we calculate the cosine similarity between the concatenated strings and the supporting text, and select as the final answer the answer with the highest similarity score.

As our transformer baseline, we use UnifiedQA, a generative model that has been trained on various types of QA tasks, including extractive, abstractive, MCQA, and yes/no questions. UnifiedQA has displayed impressive generalization capabilities, outperforming models specifically trained for indiviaul QA tasks \cite{unifiedqa}. We evaluated two versions of this model, namely base and large, and two types of inputs: one where only the question and alternatives are presented, and another in which the supporting text is also included in the input.
All models were fine-tuned for 40 epochs on the Pirá 2.0 training set with early stopping based on the development set's loss. A learning rate of 2e-5 was used for all models, and we set the batch size to the maximum available size for each model. 
We fixed a MSL of 512 for the input and 128 for the output. 

As a final baseline, we report the result of applying GPT3-turbo and GPT4 to the MCQA benchmark, under the same two kinds of input (with and without supporting text). The templates used for the prompts are found in Appendix \ref{sec:appendix}.

\subsection{Results}\label{mcqa_results}

Table \ref{tab:performanceMCQA} displays the performance of the various methods in the MCQA benchmark, where accuracy represents the percentage of correct alternatives selected in the test set. Since UnifiedQA is a text-to-text model, it cannot directly choose an alternative. Instead, it generates a text that is compared to the ground truth answer. We used the same implementation of the original UnifiedQA paper, where the alternative with the highest F1-score with the generated answer is chosen as the final answer (see Section \ref{sec:no_context}).

\begin{table}[htp]
\caption{Accuracy (in percentage) of random, IR, UnifiedQA, and GPT baselines in the MCQA benchmark. ``Supporting text'' indicates whether the supporting text was used in the prompt together with the question. Best results in bold.}
\label{tab:performanceMCQA}
\vspace*{3ex}
\centering
\setlength{\tabcolsep}{8pt}
\begin{tabular}{l|c|c}
\hline
\textbf{Model} & \textbf{Supporting text} & \textbf{Accuracy} \\ \hline
Random & - & 20\\ 
IR & - & 25.80\\
\hline
UnifiedQA base & No & 85.90 \\
UnifiedQA large & No &  87.22 \\
UnifiedQA large & Yes & 93.39 \\
\textbf{UnifiedQA base} & Yes & \textbf{95.15} \\ \hline
GPT3-turbo & No & 84.14\\
GPT4 & No & 88.55\\ 
GPT3-turbo & Yes & 94.71 \\
\textbf{GPT4} & Yes & \textbf{95.15}\\ \hline
\end{tabular}
\end{table}



The IR method only slightly outperformed random selection, implying that the Pirá 2.0 MCQA dataset cannot be resolved by identifying simple patterns of word co-occurrence. Therefore, the generation method employed to produce this dataset was effective in generating a challenging MCQA dataset.



UnifiedQA and GPT achieved a minimum accuracy of 85.90\% and 84.14\%, respectively. In both cases, the supporting text provided considerable assistance to the models as their prior knowledge was insufficient to answer some questions. 
GPT4 outperformed GPT3-turbo in both scenarios, with and without the supporting text. Interestingly, when only the question was provided, UnifiedQA large surpasssed UnifiedQA base, whereas the base version outperformed the larger version when the supporting text was included. Our hypothesis is that the larger version's knowledge was beneficial in identifying the correct alternative when no direct information was given. However, it was also more prone to distractions when the explicit context was available.

In the end, UnifiedQA and GPT demonstrated comparable results in the MCQA benchmark. When no supporting text was provided, UnifiedQA large achieved an accuracy of 87.22\%, while GPT4 achieved 88.55\%. When the supporting text was provided, both models achieved an accuracy of 95.15\%. Although less versatile than GPT3-turbo and GPT4, UnifiedQA's specialized training for QA tasks and fine-tuning for ocean-related knowledge on Pirá 2.0 led to a comparable performance in the MCQA benchmark.

\section{Discussion}\label{sec:discussion}
The two LLMs tested across the benchmarks, GPT3-turbo and GPT4, delivered exceptional results. What is more important, they did not require any training. These models were able to learn the tasks on the fly through prompt instructions. 

Although GPT3-turbo and GPT4 outperformed the models fine-tuned in the AT benchmark (Section \ref{sec:AT}), it is worth mentioning that these LLMs were not always superior to the models fine-tuned on Pirá 2.0. In the MCQA benchmark (Section \ref{sec:mcqa}), UnifiedQA matched the performance of GPT4 and surpassed that of GPT3-turbo.\footnote{The comparison is less straightforward for MRC (Section \ref{sec:mrc}) because GPT3-turbo and GPT4 did not attempt to answer some of the questions. Whether this behavior is appropriate or not depends on the specific practical needs of the task at hand.}

When choosing between different models, it is crucial to consider the tradeoff between model size and accessibility. Although LLMs eliminate the need for model training and can quickly adapt to new tasks and knowledge domains, they require significant computing power to run, making them impractical for some use cases. In contrast, smaller models such as BERT or RoBERTa can be trained on accessible GPUs like those provided by GoogleColab and can run inferences on CPUs. Table \ref{tab:models_size} provides an idea of the differences in model size.

\begin{table}[htp]
\caption{Number of parameters of the different models used in the paper.}
\label{tab:models_size}
\vspace*{3ex}
\centering
\begin{tabular}{l|l|c}
\hline
\textbf{Model} & \textbf{Version} & \textbf{\#Parameters}\\ \hline
BERT & base & 110M\\
BERTimbau & base & 110M\\
RoBERTa & base & 123M\\
PTT5 & base & 220M \\
T5 & base & 220M \\
UnifiedQA & base & 220M\\
BERT & large & 345M\\
RoBERTa & large & 354M\\
mT5 & base & 580M\\
T5 & large & 770M\\ 
UnifiedQA & large & 770M\\\hline
\multicolumn{2}{l|}{GPT3-turbo} & 175B\\
\multicolumn{2}{l|}{GPT4} & ?\\
\hline
\end{tabular}
\end{table}

Ultimately, the choice of which model to employ depends on the user's specific needs. While large language models (LLMs) offer impressive performance without the need for training, their utilization may not be justified in every scenario due to increased computational demands and potential accessibility challenges. Conversely, smaller models can be fine-tuned, a crucial feature for certain domains, such as Pirá 2.0, which is currently unavailable for GPT3-turbo and GPT4. Furthermore, it is important to note that the top-performing LLMs may come with substantial costs depending on the intended application.

\section{Limitations}\label{sec:limitations}

The effort invested in constructing these benchmarks has highlighted both the limitations of our study and avenues for future research. Our primary focus was to establish reasonable baselines, rather than to achieve state-of-the-art results, so we did not fully explore the difficulties of Pirá 2.0. Therefore, in the future, we plan to investigate solutions to overcome the challenges presented by the benchmarks.


It should be noted that while the Pirá 2.0 dataset is challenging, it is relatively small, and the provided paraphrases do not offer significant additional variation. Consequently, it is necessary to exercise caution when generalizing our results to other datasets or contexts. To address this issue, we recommend augmenting the Pirá 2.0 dataset with larger datasets such as NaturalQuestions and SQuAD. This would help to increase the overall amount of data available for training and testing.

Furthermore, we suggest that future research should explore the combination of small models and LLMs to answer questions in Pirá 2.0. For example, GPT could transform a user's question into a query, followed by a retriever model (e.g., BM25) selecting relevant texts. Then, an answerer model (e.g., T5 small), fine-tuned on oceanic domain, could be used to process the selected texts, and the resulting answer could be passed to GPT to produce a user-friendly response that takes into account the previous context. This approach may be particularly useful in ensuring that answers are both natural and accurate.

\section{Conclusion}\label{sec:conclusion}

We conducted an extensive analysis of Pirá 2.0, a machine reading comprehension dataset that focuses on the Brazilian coast, the ocean, and climate change, and proposed six benchmarks to assess its potential uses. These benchmarks covered tasks such as closed generative question answering, machine reading comprehension, information retrieval, open question answering, answer triggering, and multiple choice question answering. To facilitate the use of these benchmarks, we developed a number of baselines, including human (when available), random, and machine learning models, giving preference to solutions most commonly used in the corresponding tasks and cutting-edge large language models.

As another significant contribution, we released Pirá 2.0, a curated version of the dataset that addressed some issues in the original dataset, such as grammar errors and repeated entries. We also added several resources, including translations of supporting texts into Portuguese, answerability classification labels, automatic paraphrases of questions and answers in both languages, and multiple choice candidates, which enabled us to conduct the proposed benchmarks on Pirá 2.0. All codes are available on GitHub, and a leaderboard has been created to track future achievements in the benchmarks.

Results on the Pirá 2.0 benchmarks were generally lower than those for similar machine reading comprehension datasets, such as SQuAD 1.1 \cite{rajpurkar2016squad}, SQuAD 2.0 \cite{rajpurkar2018know}, and Natural Questions \cite{kwiatkowski2019natural}. The smaller size of Pirá 2.0, which only contains 2258 sets of questions, may partially account for this gap compared to larger resources (see Table \ref{tab:human_baselines}). However, we believe that the complex nature of Pirá 2.0, with supporting texts from a scientific domain and longer questions and answers, also contributed to the lower results. 
Although LLMs achieved a good performance on the AT and MCQA benchmarks, they demonstrated considerable difficulty on MRC. Smaller models also scored lower than humans in this task. There remains a great margin for improvement in the IR and OCQA benchmarks as well. More importantly, Pirá 2.0 stands as a unique resource for dealing with the domains of ocean, Brazilian coast, and climate change.
As a result, Pirá 2.0 has the potential to be a valuable resource for testing NLP models that use closed-domain scientific knowledge in English; for evaluating approches for machine reading comprehension in Portuguese, a language with limited resources; and as a bilingual dataset where the knowledge base is initially limited to one of the languages.



\section{Acknowledgements}
The work was carried out at the Center for Artificial Intelligence (C4AI-USP) with support from the São Paulo Research Foundation (FAPESP grant \#2019/07665-4) and from the IBM Corporation. This research was also partially supported by Itaú Unibanco S.A.; M.\ M.\ José and F.\ Nakasato have been supported by the Itaú Scholarship Program (PBI) of the Data Science Center (C2D) of the Escola Politécnica da Universidade de São Paulo.
We acknowledge support by CAPES - Finance Code 001. 
A.\ H.\ R.\ Costa and F.\ G.\ Cozman were partially supported by CNPq grants 310085/2020-9 and 305753/2022-3 respectively.
Paulo Pirozelli was supported by the FAPESP grant 2019/26762-0.

\bibliography{sn-bibliography}


\appendix
\section{Appendix}\label{sec:appendix}
Foundational models, such as GPT3-turbo and GPT4, are able to perform an incredible number of tasks. They can even execute tasks which they have never seen before, without any sort of training, by just prompting instructions.

For our benchmarks, we worked with zero-shot learning, where no examples are provided. We experimented with several versions of prompts, until finding prompts that pointed to an understanding of the task. Above, we report the prompts used in each task.\footnote{The codes for the API requests and data processing are available in \url{https://github.com/C4AI/Pira/tree/chatgpt_on_pira/ChatGPT_On_Pira}.}


\paragraph{Question Answering without Context}

\noindent {\tt EN: ``Answer in AS FEW WORDS AS POSSIBLE. ---$\backslash$n$\backslash$nQuestion: \{\}\{\}$\backslash$n$\backslash$nAnswer: ''}\\

\noindent {\tt PT: ``Responda com o MÍNIMO DE PALAVRAS POSSÍVEL. ---$\backslash$n$\backslash$nPergunta: \{\}\{\}$\backslash$n$\backslash$nResposta: ''}

\paragraph{Machine Reading Compreenhsion}

\noindent {\tt EN: ``Answer the question in AS FEW WORDS AS POSSIBLE and based on the context below.$\backslash$n$\backslash$nContext: \{\}$\backslash$n$\backslash$n\---$\backslash$n$\backslash$nQuestion: \{\}$\backslash$nAnswer: ''}\\

\noindent {\tt PT: ``Responda à pergunta com o MÍNIMO DE PALAVRAS POSSÍVEL e com base no contexto abaixo.$\backslash$n$\backslash$nContexto: \{\}$\backslash$n$\backslash$n\---$\backslash$n$\backslash$nPergunta: \{\}$\backslash$nResposta: ''}\\

\paragraph{Answer Triggering}

\noindent {\tt EN: ``This is an answer triggering task. Your task is that of telling if a question can be answered given the provided context. Your reply should be: 1: it can be answered; 0: it cannot be answered. Your reply should contain only the corresponding number and nothing else (i.e., 0 or 1). CONTEXT: \{\} QUESTION: \{\} ANSWER: ''}\\

\noindent {\tt PT: ``Este é uma tarefa de answer triggering. Sua tarefa é dizer se uma pergunta pode ser respondida com base no contexto fornecido. Sua resposta deve ser: 1: pode ser respondido; 0: não pode ser respondido. Sua resposta deve conter apenas o número correspondente e nada mais (ou seja, 0 ou 1). CONTEXTO: \{\} PERGUNTA: \{\} RESPOSTA: ''}\\

\paragraph{Multiple Choice Question Answering}

\noindent {\tt EN (no context): ``Answer the following question with the correct alternative. GIVE ONLY THE CORRECT LETTER. Question: \{\}. A: \{\}. B: \{\}. C: \{\}. D: \{\}. E: \{\}''}\\

\noindent {\tt EN (with context): ``Based on the following context: $\backslash$n$\backslash$n \{\}. Answer the following question with the correct alternative. GIVE ONLY THE CORRECT LETTER $\backslash$n Question: \{\} $\backslash$n$\backslash$n A: \{\}. B: \{\}. C: \{\}. D: \{\}. E: \{\}.}\\


\end{document}